# Cephalometric Landmark Regression with Convolutional Neural Networks on 3D Computed Tomography Data

## D. Lachinov[a,b,*], A. Getmanskaya[a,**], and V. Turlapov[a,***]

[a] Institute of Information Technologies, Mathematics and Mechanics, Lobachevsky State University
[b] Department of Ophthalmology and Optometry, Medical University of Vienna

\* e-mail: dmitry.lachinov@itmm.unn.ru
\*\* e-mail: alexandra.getmanskaya@itmm.unn.ru
\*\*\* e-mail: vadim.turlapov@itmm.unn.ru

**Abstract**—In this paper, we address the problem of automatic three-dimensional cephalometric analysis. Cephalometric analysis performed on lateral radiographs doesn't fully exploit the structure of 3D objects due to projection onto the lateral plane. With the development of three-dimensional imaging techniques such as CT, several analysis methods have been proposed that extend to the 3D case. The analysis based on these methods is invariant to rotations and translations and can describe difficult skull deformation, where 2D cephalometry has no use. In this paper, we provide a wide overview of existing approaches for cephalometric landmark regression. Moreover, we perform a series of experiments with state of the art 3D convolutional neural network (CNN) based methods for keypoint regression: direct regression with CNN, heatmap regression and Softargmax regression. For the first time, we extensively evaluate the described methods and demonstrate their effectiveness in the estimation of Frankfort Horizontal and cephalometric points locations for patients with severe skull deformations. We demonstrate that Heatmap and Softargmax regression models provide sufficient regression error for medical applications (less than 4 mm). Moreover, the Softargmax model achieves 1.15º inclination error for the Frankfort horizontal. For the fair comparison with the prior art, we also report results projected on the lateral plane.

## I. INTRODUCTION

In 1995 Richard L. Jacobson in [1] suggested applying a three-dimensional cephalometric analysis approach to orthodontic treatment. They started with the Digigraph - a device for recording soft tissue cephalometric points in 3D. Subsequent studies [2–4] show that threedimensional cephalometry is an important next step with many potential improvements over conventional two-dimensional radiographic approaches. The clear advantage of 3D analysis is the diagnosis and analysis accuracy of craniomaxillofacial deformities. This is due to the majority of congenital and acquired craniomaxillofacial deformities are asymmetric, and the deformation they correspond to is three-dimensional [5].

In turn, 3D analysis has its limitations: 3D scan study is time-consuming and perceptually hard. Moreover, Smektala et al. [6] argue that 3D cephalometry is not accurate or reproducible enough to be a reliable diagnostic technique. These flaws can be noticed while using linear and angular measurements in the traditional 2D sense. The common steps in the 3D cephalometric analysis are: 1) pick the reference frame; 2) define and annotate cephalometric points; 3) calculate the linear and angular measurements. There are reproducibility and accuracy problems at each step. The fundamental problem of choosing reproducible and convenient reference frames that are required for measuring the head and face properties is discussed in the recent studies [7–11].

According to the mentioned publications, a reference system can be classified as internal or external. The external reference system includes the midsagittal, coronal, and axial planes. They are reliable only if the head is oriented in the neutral head posture (NHP). The NHP is the position of the head where the head is neither flexed nor extended, nor rotated, nor tilted. Based on published studies, there are temporal variations in the NHP of the same patient, the intra-class variability of establishing NHP is less than 2º, an error that is not considered clinically significant. However, these studies only examined the reproducibility in pitch. The reproducibilities in roll and yaw have not yet been studied formally. If the same holds for roll, a 2º variation may cause significant problems [12].

Internal reference planes (eg, Frankfort Horizontal [FH] and sella-nasion [S-N]) are defined by internal elements such as the craniofacial bone landmarks. The reference system based on internal reference planes has several advantages. It is not affected by the head posture, doctors and researchers are familiar with these reference planes due to widespread use and normative data availability. But, the planes themselves can be distorted by craniofacial deformity or asymmetry [12]. Most studies [13–16] have suggested the construction of the reference system by appointing reference points: nasion, sella turcica, basion, orbitale, porion. Additionally, authors used crista galli, anterior nasal spine or the most superior edge of the crista galli landmarks [13-16]. However, some inner cephalometric landmarks, like Sella, are hard to identify on 3D scans [17]. At the same time, the requirements for 3D cephalometric landmarks are set in these studies. The landmarks defining the reference system have to be easy to locate, invariant to deformations and unaffectable by injuries.

Let's pay attention to the second step of cephalometric analysis. After establishing a frame we can explore landmarks' reliability. For instance, we can compare analysis by interobserver reproducibility like Raphael Olszewski et al. [18] did. They proposed a classification scheme and exclusion criteria for reference landmarks used in 3D cephalometrics that are based on inter-observer reproducibility and anatomical reality. The 3D CT cephalometric landmarks were classified into four inter-observer reproducibility groups: 1) very high, 2) high, 3) mean and 4) low. The authors demonstrated that 3D-ACRO analysis [19] was significantly more reproducible than 3D-Swennen analysis [20] due to the high number of highly reproducible cephalometric landmarks in this analysis [18].

Another problem with cephalometric analyses is fundamental. The cephalometric analysis locates points in the global reference frame but doesn't take into account facial units orientation in the space [12]. That's why some of the shape measurements might be distorted. As an example, the gonial angle can be distorted if the mandible has a roll or yaw deformity. For solving this problem Gateno et al. [12] constructed an individual coordinate system for each facial unit or element and one global "world" system for the head.

We assume that the solution to these problems can be based on the fine point-determining algorithm. The points should only depend on facial shape and have low inter-observer variability. Defining such points, we can construct robust reference systems (global and locals) and produce reliable measurements. In this study, we make a contribution to automatic annotation methods for 3D cephalometry with the intention to increase the availability of 3D analysis. In the light of recent advances in computer vision and pattern recognition, we primarily focus on deep learning for cephalometric landmark regression.

Besides the clinical application, the landmark regression is used in a number of different spheres. For instance, facial landmarks regression, human pose estimation or even crowd counting take a central part in intelligent surveillance systems. In these tasks, detected landmarks can represent different entities like face parts, body parts or whole human body. In the literature, several distinct approaches for solving the key point regression tasks can be found. The first one is the cohort of classical approaches that don't involve neural networks (NN). Other ones are related to the development of the NN. In turn, NN based methods can either be split into two groups: the direct regression of target variables and regression through some intermediate representation, for example, heatmap.

Active Appearance Models introduced by Edwards et al. [21] is an example of classical approaches. In their study, authors propose to use statistical models of shape and grey-level appearance for face landmark detection. The matching of the model to the face involves minimization of the difference between the real and synthesized face. In 1999, Chen et al. [22] used neural networks and genetic algorithms to find areas on the radiograph containing cephalometric points. Later, El-Feghi et al. [23] applied a fuzzy neural network and pattern matching method. Poyan et al. [24] used the Histograms of Oriented Gradients to describe areas that have landmarks and a support vector machine for identification. Dantone et. al. [25] proposed to use conditional regression forests for facial features detection. In contrast to the regression forest, the employed approach is conditioned on the head pose which is a global feature.

The methods based on convolutional neural networks (CNN) may be categorized into three classes. As the leader-method of first-class, we consider the direct regression method described by Osadchy et al. [26]. The authors used a convolutional neural network-based approach for mapping face image to a manifold, parametrized by pose, and non-face images to points far from that manifold. As the leader-method of second-class, we consider the heatmap based method introduced by Tompson et al. [27], were for human pose estimation used CNN, which estimates the likelihood of landmark in each spatial position as heatmap pixel values. Newell et al. [28] introduced for this task the new convolutional network heat map-based architecture named the Hourglass network. The base network operates over all scales of the image. Authors also propose to stack sequentially multiple base networks. As the third-class (Softargmax) leader, we consider Nibali et al. [29], which combine the advantages of regression and the heat map by introducing a differentiable heatmap values transformation to the spatial point coordinates.

The main part of publications, directly oriented on automatic cephalometric landmarks regression, devoted to 2D sources as Yue et. al. [30], which proposed a model-based approach locating 262 craniofacial feature points from 2D X-Ray images. For 3D sources analysis, Lachinov et al. [31] proposed to adapt the 3D template model to the pointset acquired from the patient's CT scan. For that purpose, the Coherent Point Drift algorithm [32] was used alongside a series of heuristics. Deep learning-based approaches also mainly focused on the automatic detection of cephalometric landmarks on lateral X-Ray images. Lee et al. [33] utilize patch classification and point estimation neural networks for the identification of 33 landmarks. Chen et al. [34] propose a novel attentive feature pyramid fusion module, combining it with their regression voting end-to-end trainable deep learning framework for cephalometric landmarks detection on lateral X-Ray images. Hwang et al. [35] compared the performance of the YOLOv3 [36] based system with human-annotators. The authors demonstrated that the AI-based system is as accurate as a human-annotator in the identification of cephalometric landmarks on X-Ray scans. For achieving the 3D result, Lee et al. [37] proposed the Visual Geometry Group Net-based [38] method for detecting landmarks from 2D projections of 3D CT-data. A completely 3D based approach was proposed by Kang et al. [39]. The authors develop and evaluate the 3D CNN-based system, and conclude that their 3D system couldn't achieve the accuracy needed for clinical applications, but can be used as an initial approximation for the annotators.

In each of the three classes named Direct regression, Hourglass, and Softargmax, we have chosen the most promising deep learning-based methods to adapt them to the task of three-dimensional cephalometry. Further, we will describe the methods themselves, the content of their adaptation, training and comparative study of their capabilities in solving this problem.

## II. METHODS

This section describes our approach for the cephalometric landmarks regression on the example of four main landmarks: left Orbitale; right Orbitale; left Porion; right Porion. The Orbitale (Or) is defined as the lowest point on the infraorbital margin. The Porion (Po) is defined as the most superior point located on the external auditory meatus. Each subsection corresponds to the individual deep learning model, which constructed after the search over parameters like the number of feature channels, network depth, type of base building block, feature normalization and the number of stacked networks. In this way, we got the best performing models given the inevitable limitations of the GPU memory size.

### A. Direct regression

We define a direct regression as a convolutional neural network followed by global pooling and fully connected layer. The output of the fully connected layer corresponds to the target variables. The schematic representation of the architecture is presented in Fig. 1.

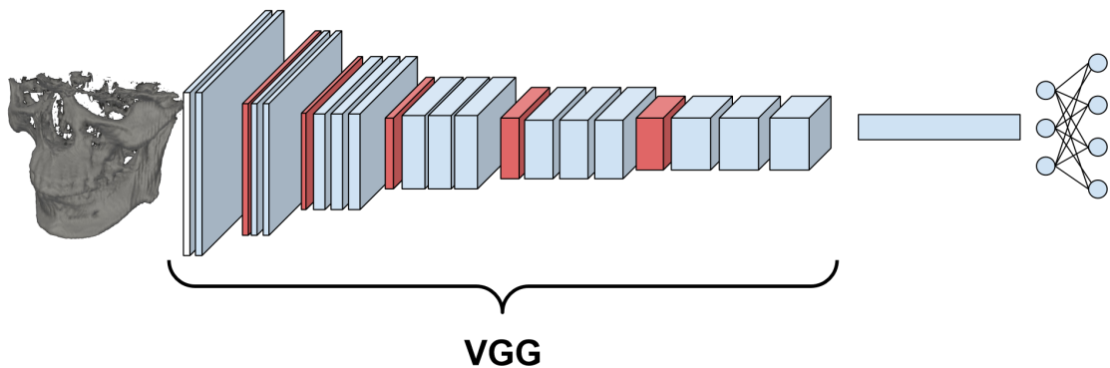

**VGG**

**Figure 1:** The VGG based architecture for direct regression. Each blue block corresponds to convolution followed by normalization and ReLU activation. The red blocks have strided convolutions.

The network follows the VGG Net (Visual Geometry Group) design introduced by Simonyan et al. [38]. It takes the full CT volume as an input and sequentially processes it with a series of blocks that consist of convolution with kernel size 3x3x3, instance normalization [40] and ReLU activation. At the end of the fully convolution part of the network global average pooling is performed that is followed by two fully connected layers with activations. The number of outputs of the final layer corresponds to the number of regressing values. In our case it equal to 4 points with 3 coordinates each, 12 in total.

### B.    Heatmap regression

In contrast to the previous model where we are trying to directly predict the target variable, here we focus on the prediction of the per-voxel likelihood of keypoint occurrence (Heatmap). The ground truth heatmaps are generated by the probability density function of Gaussian distribution with a mean in the target landmark. In the CNN design, we follow the Stacked Hourglass network architecture proposed by Newell et al. [28]. It consists of multiple subnetworks stacked one after another (Fig. 2).

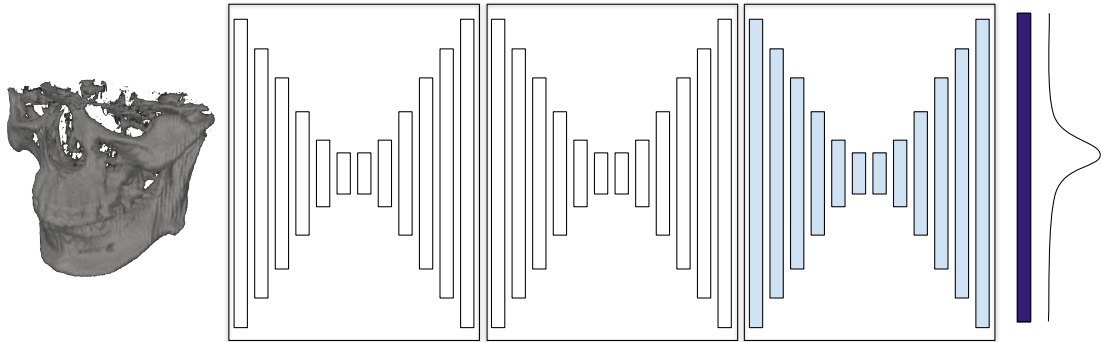

**Figure 2:** The Stacked Hourglass network consisting of stack individual Hourglass networks

The individual networks consist of encoder and decoder that are connected by the means of additive skip-connections. The architecture of an individual network is displayed in Fig. 3.

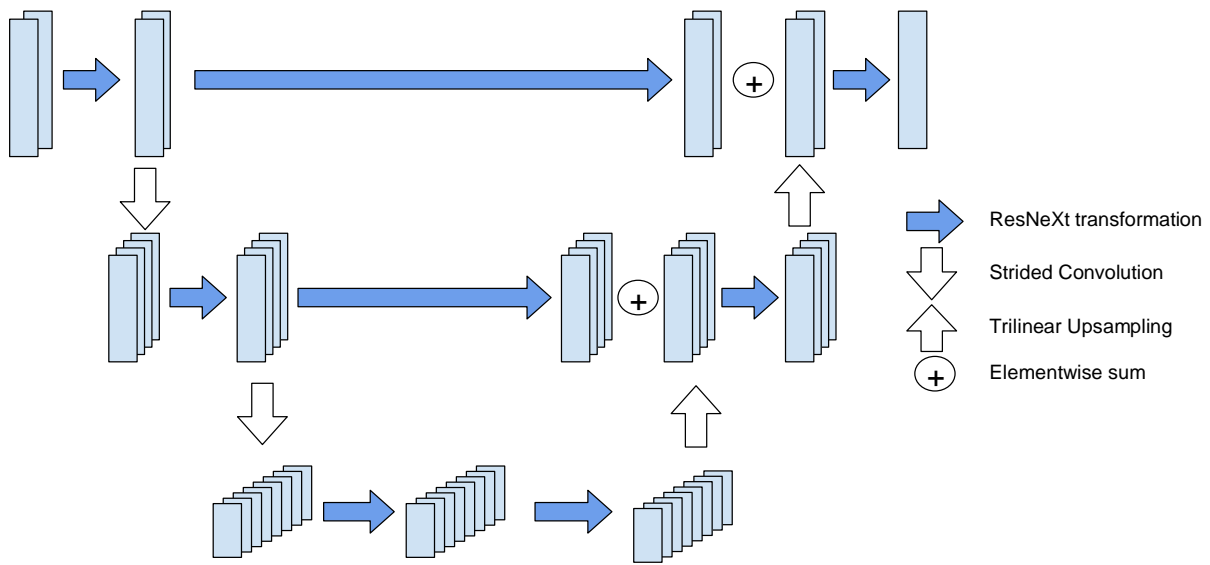

**Figure 3:** The architecture of a single hourglass network.

In this model we use 3 stacked Hourglass networks with ResNext blocks [41] and Group Normalization [42]. The output layer consists of a single convolution and sigmoid activation. At the end of each network in the stack, we provide additional supervision by attaching auxiliary output layers with the corresponding loss function.

### C. Softargmax regression

The following architecture consists of two parts: 1)The convolutional neural network; 2)Differentiable Spatial to Numerical Transform (DSNT) [29], or Softargmax [43], for transformation of the heatmap to the spatial coordinates values. Later we will refer to this transformation as a softargmax due to interpretability. In the network architecture, we use a single Hourglass network from the previous section as a backbone. Then we apply softargmax transform to its output. First, 3D feature map $M$ is normalized so that all of it's elements are non-negative and their sum is equal to one. For this, we use the rectified linear unit. After that, we normalize the feature map by dividing it by its sum:

$$\forall i, j, k \in S : M(i, j, k) \geq 0, \quad \sum_{(i,j,k) \in S} M(i, j, k) = 1,$$

where $S$ is the set of all indices. The normalized feature map can be treated as a discretized probability density function. By computing the mean of the marginal density function we calculate the target variable $y$:

$$y = (y_i, y_j, y_k), \quad y_i = \sum_{i \in S_i} i M_I(i) = S_i \cdot M_I$$

where $S_i$ is the set of indices in the $i$-th dimension, $M_I$ is the marginal density function $M_I(i) = \sum_{(j,k) \in S_{j,k}} M(i, j, k)$.

## III. EXPERIMENTS

### A. Data

Our experiment dataset consists of 20 multispiral CT images. The main feature of this dataset is that it consists of CT images of patients with significant craniomaxillofacial deformities, before and after reconstructive surgery. These real circumstances significantly increase the complexity of the Cephalometric Landmark Regression task. The resolution of each image is 800x800x496 with voxel spacing of 0.2x0.2x0.2 mm. For every image 4 cephalometric landmarks were annotated: left and right Orbitale, and left and right Porion ($Or_{left}$, $Or_{right}$, $Po_{left}$, $Po_{right}$). We utilzed the Medical Imaging Interaction Toolkit (MITK) software [44] as an annotation tool.

### B. Preprocessing

As a preprocessing step, we downsample images to the size of 128x128x64 with voxel spacing of 1.25x1.25x1.55 mm. Then we perform the z-score normalization of the image $I$ by subtracting mean μ and diving by standard deviation $\sigma$: $I_z = (I - \mu_I)/\sigma_I$ .

### C. Training

Due to the training data limited by 20 annotated CT-tomograms, we use 5-fold Cross-Validation alongside extensive data augmentation in all our experiments. We perform these experiments with PyTorch framework [45] on a single GPU with 11Gb available memory. During the training, we pick the one with the lowest RMSE values averaged across predicted points.

*a. Augmentations.* During the training procedure, we expand our dataset by performing augmentations via random contrast and random intensity shift methods. After this, we rotate the input image around a randomly chosen vector. Then we translate the image in a random direction and erase a random region of the image with a certain probability. Finally, we mirror the image in the coordinate plane. The ground truth annotation is also affected by geometric transformations. The transformed images then form a unique batch.

*b. Direct regression.* We train the regression model described in section IIA until the convergence that takes approximately $6 \cdot 10^3$ gradient updates. We use Adam as an optimizer with a learning rate of $10^{-3}$. The weight decay coefficient is set to $10^{-6}$. The network is trained with Mean Squared Error loss and batch size 1.

*c. Heatmap regression.* The network follows the design described in section IIB. The convergence happens after approximately $1.6 \cdot 10^4$ gradient update iterations. We use Adam as an optimizer with a learning rate set to $10^{-3}$ and weight decay set to $10^{-5}$. The network is trained with Binary Cross-Entropy loss and batch size 1. Additional auxiliary loss functions are connected to the outputs of the individual network.

*d. Softargmax regression.* The network follows the design described in section IIC. We train it for $1.6 \cdot 10^4$ gradient update iterations. We use Adam as an optimizer with a learning rate set to $10^{-3}$ and weight decay set to $10^{-5}$. The network is trained with Binary Cross-Entropy loss. In addition, deep supervision for each decoder level is employed. This is achieved by adding auxiliary outputs followed by softargmax operation and batch size 1.

## D. Evaluation

For evaluation, we perform prediction of the left and right Orbitale and Porion with 5 fold Cross Validation on the dataset of 20 images. We compute Root Mean Squared Error (RMSE) between predicted and annotated points and inclination angle between predicted and annotated Frankfort Horizontals (FH). In the evaluation, we focus on RMSE values for individual points and on total RSME for every model. Besides this, we investigate cumulative distribution plots since they play a crucial role in assessing the methods for clinical applicability. We separately report the likelihood of individual landmarks to be in the range of 2, 3 and 4 mm from the target. Finally the same analysis has been performed for the lateral projection. It provides the comparison of our 3D results with the results obtained by other methods operating on 2D datasets.

## IV. RESULTS

The RMSE values for landmark points in all our experiments are reported in Tables 1-4. If we compare RMSE values for all landmark points in the 3D (Table 1) and on the lateral projection (Table 2) we will see that all models have lower error rates except the Direct Regression. Besides, our observation about differences in Porion predictions for different models still holds in case of lateral projection.

**Table 1:** The 3D RMSE for landmarks (mm)

| Model | $Or_{left}$ | $Or_{right}$ | $Po_{left}$ | $Po_{right}$ | Total |
|---|---|---|---|---|---|
| Direct Regr. | 4.14 | 3.78 | 5.03 | 5.22 | **4.58** |
| Hourglass | 1.66 | 1.24 | 1.91 | 1.97 | **1.72** |
| Softargmax | 1.77 | 1.57 | 2.08 | 1.80 | **1.81** |

**Table 2:** The lateral RMSE for landmarks (mm)

| Model | $Or_{left}$ | $Or_{right}$ | $Po_{left}$ | $Po_{right}$ | Total |
|---|---|---|---|---|---|
| Direct Regr. | 3.71 | 3.91 | 5.31 | 5.75 | **4.75** |
| Hourglass | 0.99 | 1.03 | 1.69 | 1.97 | **1.44** |
| Softargmax | 1.23 | 1.27 | 2.01 | 1.77 | **1.61** |

The average likelihood of falling cephalometric landmarks within the predefined radius via CNN-regression is reported in Table 3 for the 3D case, and in Table 4 for the lateral projection only. The error statistics shown of Tables 1,3 are also analyzed in detail in the likelihood curves form (Fig. 4) and in the form of the box-and-whiskers diagrams or box plots (Fig. 5).

**Table 3:** The 3D likelihood

| Model | 2 mm | 3 mm | 4 mm |
|---|---|---|---|
| Direct Regr. | 0.20 | 0.39 | 0.61 |
| Hourglass | 0.81 | 0.94 | 0.95 |
| Softargmax | 0.78 | 0.92 | 0.97 |

**Table 4:** The lateral likelihood

| Model | 2 mm | 3 mm | 4 mm |
|---|---|---|---|
| Direct Regr. | 0.21 | 0.48 | 0.69 |
| Hourglass | 0.89 | 0.97 | 0.97 |
| Softargmax | 0.88 | 0.97 | 0.97 |

Figure 4 shows the cumulative plots of Likelihood vs Error in 3D space (mm) for all three CNN investigated models. Figure 4a provides an average likelihood of the CNN models. The error value 4 mm is marked by red thick vertical because values down 4 mm are considered applicable for clinical use. As we can see here, for Direct Regression CNN only 61% of the predicted points fall within the 4 mm radius. In contrast, Stacked Hourglass and Softargmax models achieve 95% and 97% falling within the 4 mm radius respectively. The high accuracy for 2 mm and 3 mm radius is also notable (on figures and in Table 3). In Fig. 4b we can see also that the predictions of Direct Regression distributed almost uniformly with respect to error (distance to the ground truth). In the presence of high training accuracy, this is a sign of insufficient model generalization. The prediction distributions of other models behave identically (Fig. 4c and 4d) reaching high accuracy values at the 3 mm threshold.

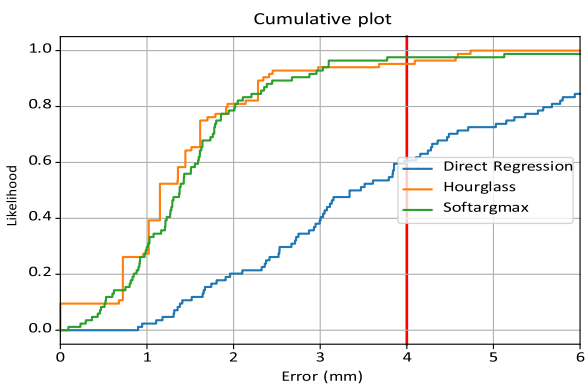

(a) Averige likelihood for all three CNN models

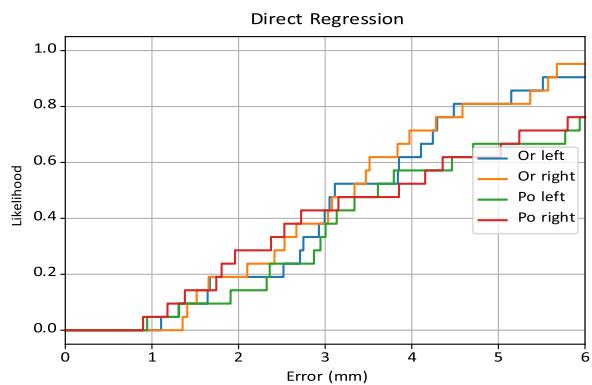

(b) Likelihood for landmarks via Direct Regression

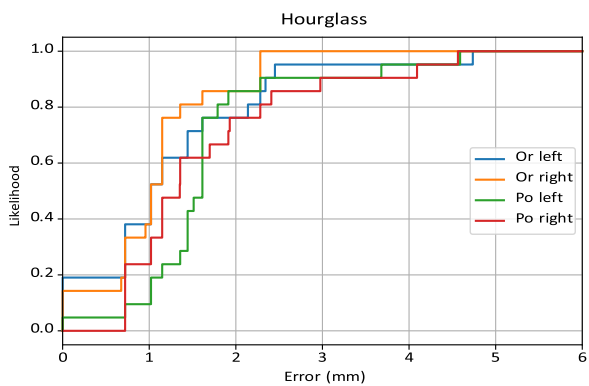

(c) Likelihood for landmarks via Hourglass

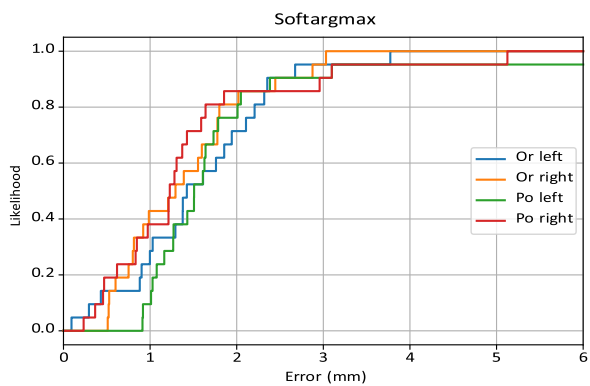

(d) Likelihood for landmarks via Softargmax

**Figure 4:** The cumulative plots of Likelihood vs Error (mm)

The Porion landmarks have higher average error compared to Orbitale for each CNN-model (Fig. 4b-d). And really, the Porion landmarks are harder to annotate than the Orbitale.

Fig.5 shows the error probability distribution for cephalometric landmarks detection via three types of CNN in the form of box-and-whiskers diagrams. Fig. 5a demonstrates that Direct Regression has the highest error among all three methods. Other figures show curves of Likelihood vs Error (in mm) for the landmarks detection via Direct Regression (Fig. 5b), Hourglass (Fig. 5c) and Softargmax (Fig. 5d) networks separately. Comparing box plots for the Stacked Hourglass model (Fig. 5c) and Softargmax based model (Fig. 5d), we see that despite having lower average value, the Porion predictions of the Softargmax model have lower spread and fewer outliers.

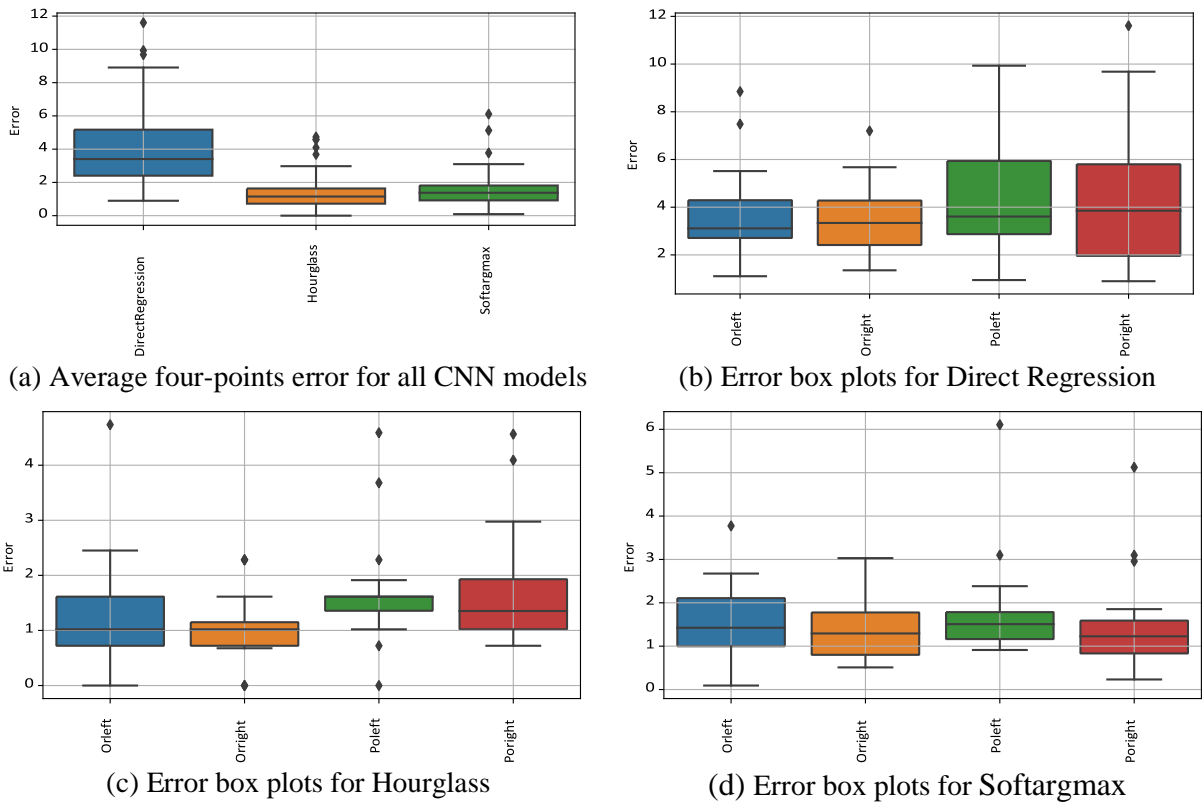

(a) Average four-points error for all CNN models

(b) Error box plots for Direct Regression

(c) Error box plots for Hourglass

(d) Error box plots for Softargmax

**Figure 5:** The error probability distribution for CNN-detection the cephalometric landmarks

Figure 6, as Figure 4, shows the cumulative plots of Likelihood vs Error (mm), but for traditional lateral skull projection only. See also Table 2 and Table 4 for RMSE of landmark predictions and the likelihood of falling within the predefined error value. The accuracy of prediction on the traditional lateral skull projection is higher than on 3D because the error for lateral projection is a component of the magnitude of the 3D error. As in the 3D case, the predictions of Direct Regression distributed almost uniformly with respect to the error (Fig. 6b). Despite having lower RMSE, the Softargmax model (Fig. 6d) reaches the same likelihood values as the Hourglass model (Fig. 6c). Both models have 97% of points that fall within the 3 mm error value. All curves of Figures 4 and 6 have stepwise shapes due to the discrete nature of the predictions determined by downsampling, the heatmap grid size and using a loss-function without regularization member. The curves for Hourglass CNN has higher step size (Figures 4c and 6c).

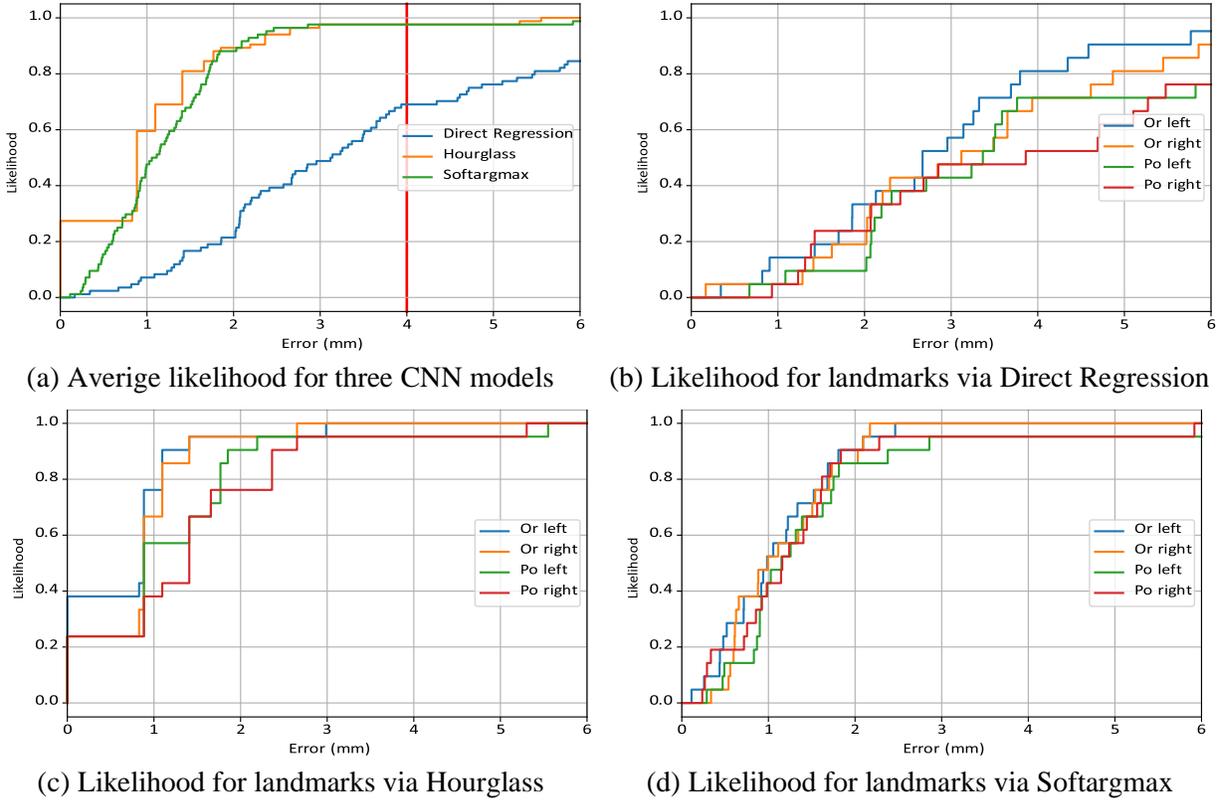

(a) Averige likelihood for three CNN models

(b) Likelihood for landmarks via Direct Regression

(c) Likelihood for landmarks via Hourglass

(d) Likelihood for landmarks via Softargmax

**Figure 6:** The cumulative plots of Likelihood vs Error (mm) for lateral projection

The four main reference points ($Or_{left}$, $Or_{right}$, $Po_{left}$, $Po_{right}$) are the basis for constructing coordinate systems at which other cephalometric points and standard reference planes can be defined. The most famous and stable reference plane is the Frankfort Horizontal (FH). We have evaluated the accuracy (in terms of inclination angle value) for defining the Frankfurt Horizontal from the four reference points obtained using the three different studying CNNs (Fig. 7). First, we obtained human-estimated reference FHs and FHs by fitting the linear regression to the main landmarks. Then we calculated the inclination angle from reference FH for each plane. The results are demonstrated in Fig. 7. The average inclination error is 3.41º, 1.18º and 1.15º for Direct Regression, Hourglass, and Softargmax models respectively. For this characteristic, the Softargmax model outperforms the other two methods.

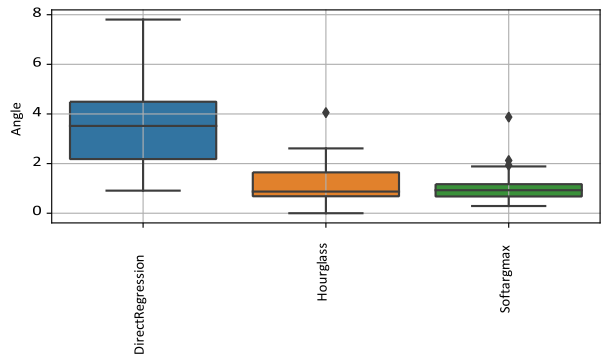

**Figure 7:** Distributions for Frankfort Horizontal inclination angle

## V.   CONCLUSION

In this study, we analyzed the current state of the convolutional neural network (CNN) for medical and cephalometry applications. We adopted and studied three specific models of CNN for cephalometric landmarks detection. The models were named as Direct regression, Heatmap regression, Softargmax regression. We investigated the possibilities of training and using these models for the regression of cephalometric points directly on 3D CT scans. For the first time, the data with significant craniomaxillofacial deformations of patients before and after corrective surgery, in the amount of 20 CT, were used for training. We employed 5-fold Cross-Validation and traditional augmentation methods for training. The training was conducted in the PyTorch framework on a single GPU with 11Gb available memory.

We demonstrated that two out of three models, Heatmap and Softargmax, provide sufficient regression error for medical applications (less than 4 mm), even in the presence of significant deformations of the skull. The best out of the two (Heatmap and Softargmax, $1.18^{\circ}$ and $1.15^{\circ}$, respectively) models w.r.t. angular deviation of the Frankfort horizontal was the Softargmax model. Finally, we analyze the results for the lateral projection. It enables the comparison of our 3D results with prior methods operating on 2D datasets. As a result, Heatmap and Softargmax regressions correctly estimate 97% of the points within 4 mm from the ground truth projection.

ACKNOWLEDGEMENTS This work was supported by the Russian Foundation for Basic Research, project no. 18-37-00383.

CONFLICTS OF INTEREST. The authors declare they have no conflicts of interest.